%% file: main.tex
\pdfoutput=1
\documentclass[10pt,twocolumn,letterpaper]{article}

\usepackage{wacv}              % To produce the CAMERA-READY version
\usepackage{graphicx}
\usepackage{amsmath}
\usepackage{amssymb}
\usepackage{booktabs}
\usepackage{multirow}
\usepackage[accsupp]{axessibility}

\usepackage[pagebackref,breaklinks,colorlinks]{hyperref}

\usepackage[capitalize]{cleveref}
\crefname{section}{Sec.}{Secs.}
\Crefname{section}{Section}{Sections}
\Crefname{table}{Table}{Tables}
\crefname{table}{Tab.}{Tabs.}

\def\wacvPaperID{438}
\def\confName{WACV}
\def\confYear{2025}

\begin{document}

\title{BIV-Priv-Seg: Locating Private Content in \\ Images Taken by People With Visual Impairments}

\author{
Yu-Yun Tseng\textsuperscript{1}, Tanusree Sharma\textsuperscript{2}, Lotus Zhang\textsuperscript{3}, Abigale Stangl\textsuperscript{4}, Leah Findlater\textsuperscript{3}, \\
Yang Wang\textsuperscript{2} and Danna Gurari\textsuperscript{1}
\\
{\small [1] University of Colorado Boulder, [2] University of Illinois at Urbana-Champaign,} \\ {\small [3] University of Washington, [4] Georgia Institute of Technology}}

\maketitle

%%%%%%%%% ABSTRACT
\begin{abstract}

Individuals who are blind or have low vision (BLV) are at a heightened risk of sharing private information if they share photographs they have taken.  To facilitate developing technologies that can help them preserve privacy, we introduce BIV-Priv-Seg, the first localization dataset originating from people with visual impairments that shows private content.  It contains 1,028 images with segmentation annotations for 16 private object categories. We first characterize BIV-Priv-Seg and then evaluate modern models' performance for locating private content in the dataset.  We find modern models struggle most with locating private objects that are not salient, small, and lack text as well as recognizing when private content is absent from an image. We facilitate future extensions by sharing our new dataset with the evaluation server at 
\texttt{https://vizwiz.org/tasks-and-datasets/o\\bject-localization/}.
\end{abstract}

\input{01-Introduction}
\input{02-Related_Work}
\input{03-Dataset}

\input{04-Benchmarking}
\input{05-Conclusion}

\paragraph{\bf Acknowledgments.}
This project was supported by National Science Foundation SaTC awards (\#2148080, \#2126314, and \#2125925).

% \bibliography{egbib}

%%%%%%%%% REFERENCES
{\small
\bibliographystyle{ieee_fullname}
\bibliography{references}
}

\end{document}

%% file: 01-Introduction.tex
\section{Introduction}
\label{sec:introduction}
People who are blind or have low vision (BLV) regularly share photographs they have taken.  They do so for multiple reasons including to stay connected on social media~\cite{ahmed2015privacy,bennett2018teens,voykinska2016blind} and to receive assistance with everyday visual tasks~\cite{BeMyEyes:online,SeeingAI:online}, such as recognizing objects and reading mail.  

\emph{A concern is that many images from BLV photographers contain private content~\cite{ahmed2015privacy}}.  Underscoring this concern, one study showed that privacy leaks occurred for over 10\% of more than 40,000 images shared by BLV photographers using a visual assistance technology (i.e., the VizWiz application)~\cite{gurari2019vizwiz}.  Such privacy leaks are concerning regardless of whether they are seen by remote humans providing visual assistance or retained by companies~\cite{stangl2023dump,stangl2022privacy}, who may subsequently use such data for training AI models and for accruing licensing revenue from other end users.

\begin{figure}[t!]
\centering
	\includegraphics[width=\linewidth]{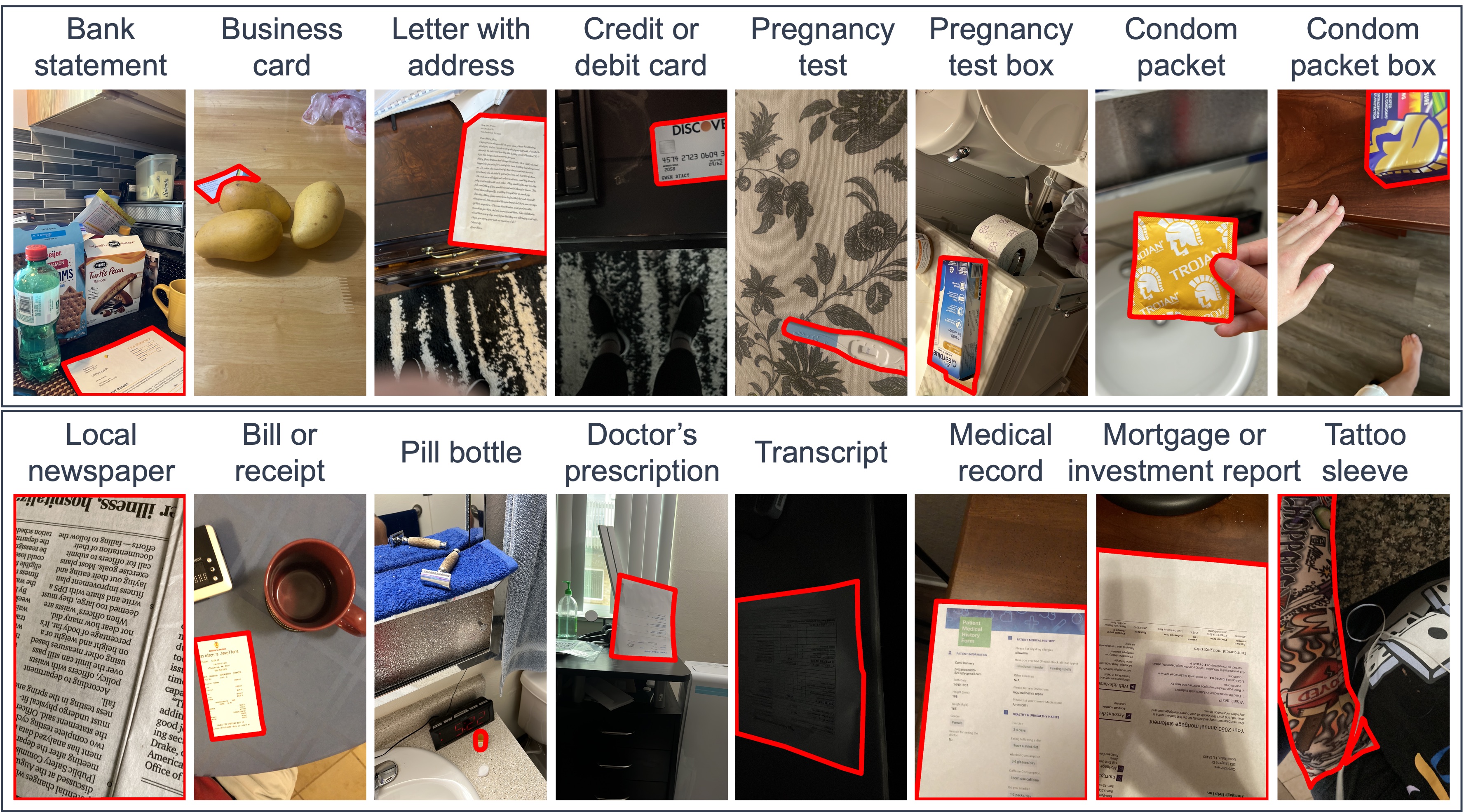} \hfill 
 \vspace{-1.5em}
  \caption{We introduce BIV-Priv-Seg, a dataset containing segmented private objects. An example is shown for each category.}
  \label{fig:motivation}
\end{figure}

To facilitate the development of privacy-preserving technologies, we introduce the first localization dataset that originates from people with visual impairments and shows private content. Called \textbf{BIV-Priv-Seg}, it extends prior work's~\cite{sharma2023disability} collection of images that were taken by BLV photographers of (non-personal) private content. For each of the 1,028 images in the dataset, we collected human-annotated segmentations for every instance of the 16 photographed private object categories. This culminated in $967$ instance segmentations. Examples of annotated images from every private object category are shown in \textbf{Figure~\ref{fig:motivation}}.  

We also characterize BIV-Priv-Seg and compare it with five existing few-shot localization datasets to reveal what makes it similar to prior work as well as unique. Similarities to prior findings about images from blind photographers include that objects exhibit greater variability in terms of how much of an image they occupy and objects reside more often on image borders. Differentiating aspects of BIV-Priv-Seg include the lack of presence of target objects in 19\% of the images, a high rate of images containing one annotated object, and a higher prevalence of objects containing text. 

Finally, we benchmark three modern few-shot localization models and vision-language models to reveal their current strengths as well as where improvements are needed. Our fine-grained analysis reveals they struggle most at locating objects that are not salient, small, and lack text as well as recognizing when private content is absent from an image. We expect this work will facilitate the development of models that can handle a greater diversity of real-world challenges and also benefit a larger audience encountering similar privacy challenges, such as children and individuals with cognitive impairments.

%% file: 02-Related_Work.tex
\section{Related Work}
\label{sec:related}

\paragraph{Computer Vision Datasets.}
A critical component for successfully developing computer vision models is the availability of human-annotated datasets to support reproducible, quantitative evaluation.  A challenge for developing privacy-aware technologies is that most datasets lack private content due to intentional filtering efforts, including popular, mainstream datasets such as VOC~\cite{dataset_voc}, ImageNet/ILSVRC~\cite{russakovsky2015imagenet}, COCO~\cite{dataset_coco}, LSUN~\cite{yu2015lsun}, and Open Images~\cite{kuznetsova2018open}. Only relatively small collections of private images (e.g., thousands of examples) have been successfully curated for public use, including by scraping them from photo-sharing websites (i.e., Flickr, Twitter)~\cite{orekondy2018connecting,orekondy2017visualprivacyadvisor,zerr2012Privacyawareimageclassification} and hiring BLV photographers to photograph (non-personal) private content~\cite{sharma2023disability}.  Extending the latter prior work, \emph{we introduce the first publicly-available dataset originating from BLV photographers that shows segmented private content}.  This effort extends the broader research community's inclusivity efforts to increase the availability of datasets based on images taken by people with vision impairments~\cite{Bafghi_2023_CVPR, bhattacharya2019does,chen2022grounding,chiu2020assessing,dataset_VizWiz,dataset_VizWizCaption,dataset_KVAQ,dataset_orbit,gurari2018predicting,gurari2019vizwiz,zeng2020vision} while following the standard practice of augmenting annotations to images that are already used in other datasets~\cite{tseng2022vizwiz,gupta2019lvis,fsod_attenRPN,dataset_coco}. 
 
Given that it is non-trivial to curate a large-scale collection of private images for supporting the development of privacy-aware models, since the very nature of such data means it typically cannot be saved or shared, we focus on introducing a dataset challenge that facilitates the development of zero-shot and few-shot models (and so requires a small amount of training data).  Accordingly, our dataset challenge will complement prior work that proposed few-shot object detection and instance segmentation challenges, including VizWiz-FewShot~\cite{tseng2022vizwiz}, PASCAL-$5^{i}$~\cite{fsss_oneshot}, COCO-$20^{i}$~\cite{fsss_FeatureWeighting,fsis_siameseMRCNN}, LVIS~\cite{fsis_ifs-RCNN}, and FSOD~\cite{fsod_attenRPN}.  Our dataset offers unique challenges compared to existing datasets, including images lacking the target object, a wider variability of object sizes, a higher prevalence of objects residing on image borders, and a higher prevalence of objects with text. 

\vspace{-1em}\paragraph{Zero-Shot and Few-Shot Object Localization Models.}
Zero-shot and few-shot methods that can locate objects in images span more traditional few-shot object localization models and more recent vision-language models.  Traditional methods include those designed to only achieve strong performance on novel object categories~\cite{fsod_DeFRCN,fsis_siameseMRCNN,fsod_defrcn_aug, fsod_NormVAE,fsod_DandR,fsod_ICPE,fsod_VFA,fsod_sigma-adaptive,fsis_FGN,fsis_FAPIS,fsis_ifs-RCNN} as well as those 
designed to also maintain performance on base object categories~\cite{fsod_DETRreg, fsod_NIFF} (i.e., generalized/incremental setting). A new breed of relevant models also emerged in the past year: large multimodal models (LMMs) that can generate a textual description alongside a visual grounding (i.e., segmentation mask) specifying where the described content is located~\cite{hanoona2023GLaMM,wang2023cogvlm}.  LMMs offer the advantage of being more interactive, making them particularly appealing for assistive tools for the BLV community.  We benchmark a total of three models (2 traditional and 1 LMM) to assess how well modern methods perform on our new dataset.  While we observe the best performance from LMMs, all models still fall below human performance.  Our results suggest that promising future directions for research are to improve models' abilities to handle objects that are not salient, small, and lack text as well as to recognize when private content is absent from an image. 

%% file: 03-Dataset.tex
\section{BIV-Priv-Seg Dataset}
\label{sec:dataset}
We now introduce our dataset for locating private objects in images taken by people with vision impairments.

\subsection{Dataset Creation}

\paragraph{Data Source.}
We leverage the BIV-Priv dataset~\cite{sharma2023disability}, which consists of 1,176 images taken by 26 BLV photographers of 16 types of private objects.  All private objects were `props', meaning that the objects lacked any private information pertinent to the photographer (e.g., a fake medical bill rather than a real one).  Each BLV individual took pictures with each object type positioned to appear in the image's foreground as well as its background. We leverage all images, including those that lacked the target private content, to reflect the unique challenges faced by BLV photographers who often do not successfully capture the content of interest in their images.

\vspace{-1em}\paragraph{Category Selection.}
After slightly modifying the categories in BIV-Priv~\cite{sharma2023disability}, as discussed in the Supplementary Materials, we chose $16$ private object categories as follows: ``bank statement", ``bill or receipt", ``business card", ``condom box", ``condom packet", ``credit or debit card", ``doctor's prescription", ``pill bottle", ``letter with address", ``local newspaper", ``medical record document", ``mortgage or investment report", ``pregnancy test", ``pregnancy test box", ``tattoo sleeve", and ``transcript".

\vspace{-1em}\paragraph{Data Filtering.}
We filtered all images containing any personally identifiable information (PII) from the photographer in a multi-step approach.  After receiving approval from Institutional Review Boards (IRB), two authors who were trained to recognize all `prop' private objects and the definition of PII reviewed every image independently. When both individuals flagged an image as containing PII, it was filtered. When only one flagged an image, it was discussed by three authors to make a final decision. This process yielded $1,028$ images suitable for public release. 

\vspace{-1em}\paragraph{Annotation Task.}
We leveraged the instructions and annotation protocol established by prior work (i.e., VizWiz-FewShot~\cite{tseng2022vizwiz}) to enable fair comparison with it.  To create instance segmentations, annotators clicked a series of points to generate polygons, dragged a polygon's vertices to refine it, and created holes by drawing `interior' polygons in `exterior' polygons.  We leveraged the annotation platform, \emph{Supervise.Ly} (i.e., https://supervisely.com/) to collect all segmentation annotations.

\vspace{-1em}\paragraph{Annotation Collection.}
We next localized all instances of the 16 private object categories in every image.  To collect high-quality annotations, we relied on three in-house annotators who were authors of this work and so intimately familiar with the images (matching the approach employed for creating ADE20K~\cite{zhou2019semantic}).  Each individual had a 1-hour 1-on-1 training session to learn about the annotation interface and then passed a test demonstrating they could accurately annotate a sample image.  Each image was annotated by two annotators and we used $IoU$ scores to determine how to establish a ground truth segmentation per image (matching the approach employed for COCO-Stuff~\cite{caesar2018coco}).  When $IoU \geq 0.8$, we randomly chose one annotation for the ground truth.  Otherwise, the three authors reviewed the pair of annotations to choose one as the ground truth (or, in exceptional cases, discarded both annotations). Only $15\%$ of the images required manual review, with $93\%$ having one of the annotations deemed correct. This effort culminated in 967 instance segmentations.

% \begin{figure}[b!]
% \centering
% 	% \includegraphics[width=1.0\linewidth]{images/unique_features.png} \hfill 
% 	\includegraphics[width=1.0\linewidth]{images/unique_features.pdf} \hfill 
%     \vspace{-1.5em}
%   \caption{Examples from our BIV-Priv-Seg dataset illustrating its unique aspects, specifically a high variability for the proportion of images occupied by the objects (with many objects occupying much of their images' borders), high prevalence of text in objects, and inclusion of images lacking target objects.}
%   \label{fig:BIV-unique-features}
% \end{figure}

\begin{figure}[b!]
\centering
	\includegraphics[width=1.0\linewidth]{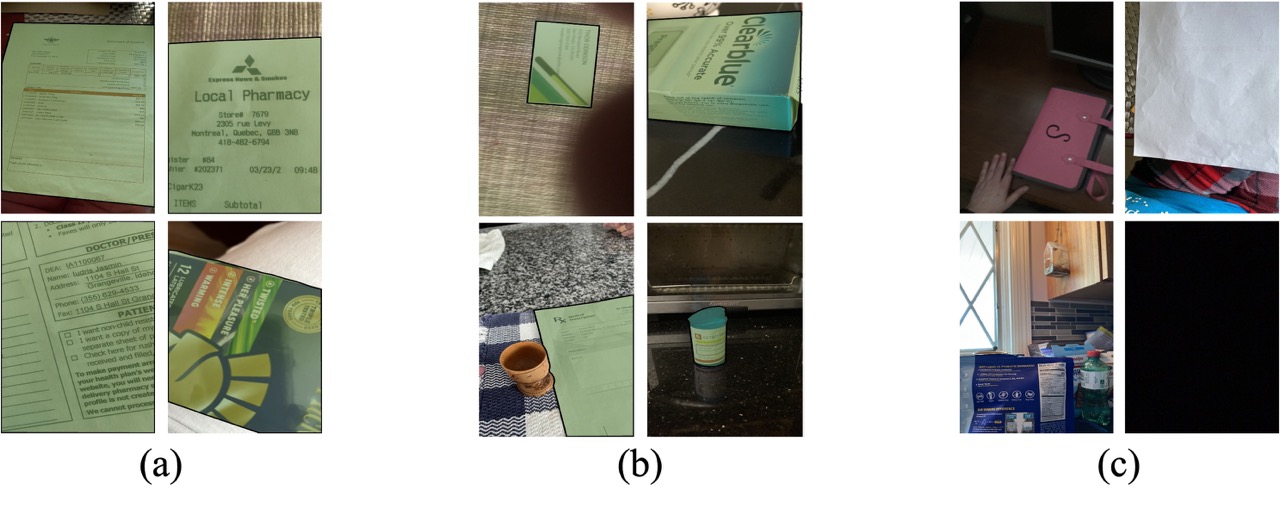} \hfill 
    \vspace{-1.8em}
  \caption{Examples from BIV-Priv-Seg dataset illustrating its unique aspects: (a) high proportion of image areas and borders occupied by the objects, (b) high prevalence of objects containing text, and (c) images lacking target objects.}
  \label{fig:BIV-unique-features}
\end{figure}

\subsection{Dataset Analysis}
We now characterize BIV-Priv-Seg and how it compares to the following mainstream few-shot localization datasets: VizWiz-FewShot~\cite{tseng2022vizwiz}, PASCAL-5\textsuperscript{i}~\cite{fsss_oneshot}, COCO-20\textsuperscript{i}~\cite{fsss_FeatureWeighting}, LVIS~\cite{fsis_ifs-RCNN} and FSOD~\cite{fsod_attenRPN}.  Our findings will show that our dataset offers several unique aspects compared to existing datasets, which are exemplified in \textbf{Figure~\ref{fig:BIV-unique-features}}.

\begin{figure*}[t!]
\includegraphics[width=\textwidth]{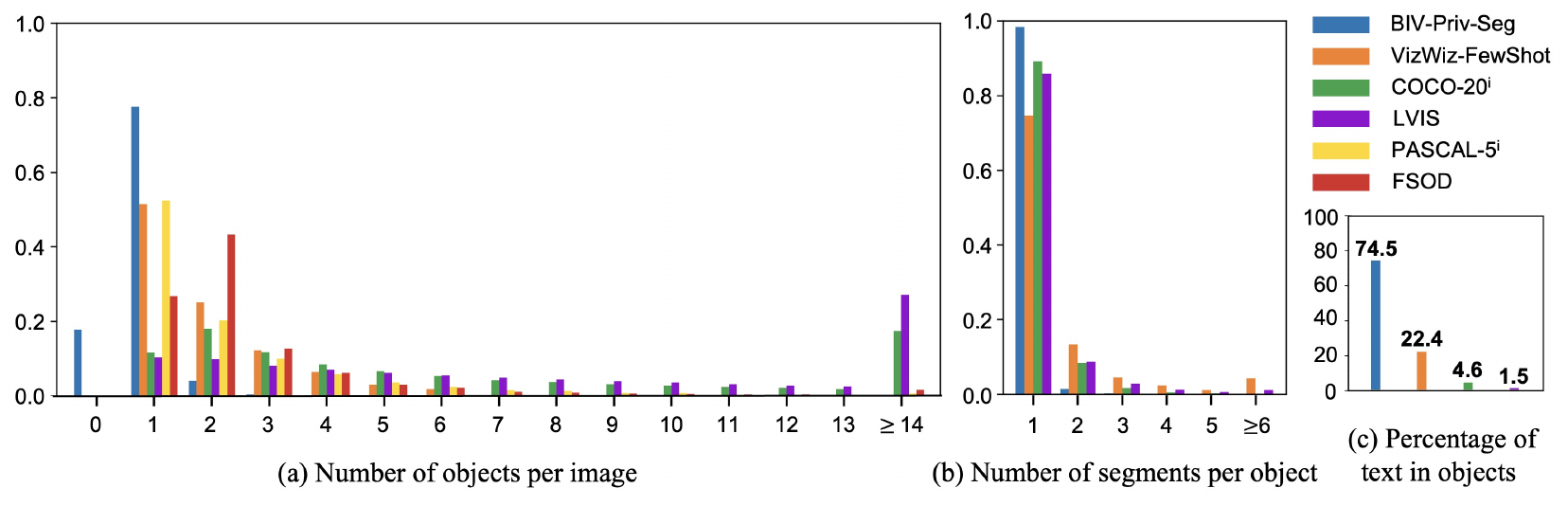}
\vspace{-1.8em}
\caption{Comparison of our dataset to four existing few-shot localization datasets with respect to the (a) number of annotated objects per image, (b) number of segments (i.e., disconnected areas) per annotated object, and (c) percentage of annotated objects containing text.}
\label{fig:global_analysis}
\end{figure*}

\vspace{-1em}\paragraph{Global Dataset Statistics.}
We analyze three properties for each few-shot localization dataset: the number of annotated objects per image, the number of disconnected areas an object contains (i.e., \emph{segments}), and the percentage of objects containing text\footnote{Objects are flagged as containing text if Microsoft Azure's Optical Character Recognition (OCR) API detects text within the image, after masking out all content except for the instance segmentation.}. PASCAL-5\textsuperscript{i} and FSOD are excluded from the latter two metrics since their annotations only contain bounding boxes and so lack the necessary segmentation information to count segments and locate texts. Results are visualized in \textbf{Figure~\ref{fig:global_analysis}}.

\begin{figure*}[t!]
\includegraphics[width=1.0\textwidth]{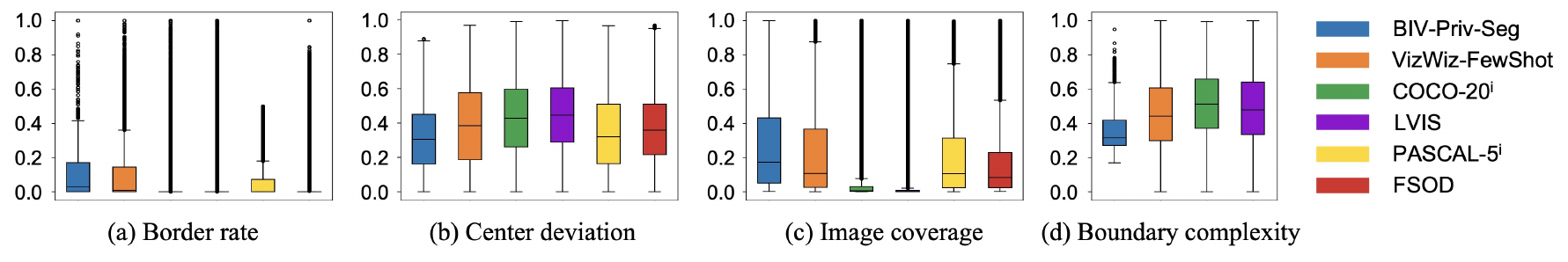}
\vspace{-1.8em}
\caption{Comparison of our dataset to four existing few-shot localization datasets with respect to the (a) border rate, (b) center deviation, (c) image coverage, and (d) boundary complexity. PASCAL-5\textsuperscript{i} and FSOD are excluded for boundary complexity as they lack the necessary segmentation annotations to calculate this metric.}
\label{fig:localization_analysis}
\end{figure*}

From the histogram showing the \emph{percentage of images containing each number of annotated objects} (\textbf{Figure~\ref{fig:global_analysis}(a)}), we observe two key distinctions of our dataset.  \emph{First}, our dataset is the only one with \emph{images lacking objects of interest} with it occurring for 18\% of the images.  These reflect both when no target object is visible and when the target object is a paper positioned on its blank side, rather than the side showing the document's text (and so the private text).  This amount is similar to that reported by prior work~\cite{chiu2020assessing}, where roughly 20\% of BLV photographers' images lacked the visual evidence needed to answer their visual question despite showing recognizable content~\cite{dataset_VizWizCaption}.  We retain such images in the final dataset to reflect that the photographers \emph{intended} to capture an object of interest in such images~\cite{sharma2023disability}, but exclude these examples for subsequent data analysis to support fair dataset comparisons. Retaining these target-less samples will facilitate developing models that recognize when no target objects are present, which is relevant for many real-world applications.  

The \emph{second} key distinction of our dataset based on \textbf{Figure~\ref{fig:global_analysis}(a)} is that it rarely has more than one annotated object per image.  This trend aligns with findings from prior work~\cite{gurari2019vizwiz}, which show that only one private object is typically observed in an image when privacy leaks arise for BLV photographers authentically trying to learn about their visual surroundings.  We attribute this finding in our dataset to the distinct photography goal for BIV-Priv~\cite{sharma2023disability}, where images were intentionally photographed with a single private object.  However, we suspect many images in BIV-Priv \emph{show multiple objects}, because when evaluating the performance of a salient object detection model~\cite{saliency_kim2022revisiting} (shown to yield comparable performance to humans for images taken by people with visual impairments~\cite{reynolds2024salient}), we found that only 32\% of the annotated private objects overlap significantly (IoU $\geq 0.5$) with salient areas. The presence of multiple objects could be in part due to how images were curated~\cite{sharma2023disability}, as photographers were instructed to position private objects in the background for half the images. 

When observing the histogram showing the \emph{percentage of images containing different numbers of segments (i.e., disconnected areas)} (\textbf{Figure~\ref{fig:global_analysis}(b})), we observe that our BIV-Priv-Seg dataset typically exhibits only one segment per object. VizWiz-FewShot, in contrast, more often has the most with up to $122$ segments in an object. While an object itself exists as a unified entity in the real world, multiple counts of segments arise in an object under two circumstances: (1) when occlusions break objects into more than one region and (2) when objects are partially outside the frame of the image so there are disconnected regions of an object. Our dataset limits these scenarios, thereby permitting more focused model benchmarking analysis for other challenges that can arise, as discussed in this section.

When analyzing the \emph{prevalence of text in objects}, we find it is the highest for our dataset, occurring for $74.5\%$ of annotated object instances (\textbf{Figure~\ref{fig:global_analysis}(c)}).  We attribute this greater prevalence of text to the intrinsic nature of the private categories, where textual information often \emph{is} the private content. For instance, paper-based categories such as ``bill or receipt" and ``doctor's prescriptions" can show names, addresses, and more.  Consequently, our dataset provides an opportunity to shift algorithm developers' focus to more heavily prioritize analyzing text on objects.

\begin{table*}[t!]
  \centering
  \footnotesize
  \vspace{-0.5em}
    \begin{tabular}{lcccccccccccccccc}
\toprule
& \multicolumn{4}{c}{FSOD mAP} & \multicolumn{4}{c}{FSOD AP50}  & \multicolumn{4}{c}{FSIS mAP} & \multicolumn{4}{c}{FSIS AP50}\\
\cmidrule(r){2-5}  \cmidrule(r){6-9} \cmidrule(r){10-13} \cmidrule(r){14-17}  
Dataset~~~\textbackslash~~~$k = $ & 1 & 3 & 5 & 10 & 1 & 3 & 5 & 10 & 1 & 3 & 5 & 10 & 1 & 3 & 5 & 10\\ 
\midrule 
BIV-Priv-Seg (ours) & \textbf{18.1} & \textbf{23.3} & \textbf{28.7} & \textbf{40.0}  & \textbf{24.7} & \textbf{31.7} & \textbf{39.4} & \textbf{52.4} &  \textbf{20.0} & \textbf{25.6} & \textbf{31.1} & \textbf{44.4} & \textbf{24.5} & \textbf{31.3} & \textbf{38.5} & \textbf{52.2}\\
VizWiz-FewShot~\cite{tseng2022vizwiz} & 10.7 & 11.7 & 22.6 & 32.5 & 13.2 & 14.9 & 26.9 & 43.2 & 10.0 & 10.5 & 20.3 & 28.7 & 12.0 & 12.8 & 25.5 & 36.4 \\
COCO-20\textsuperscript{i}~\cite{fsss_FeatureWeighting} & - & - & - & - & 19.0 & - & 21.7 & - & - & - & - & - & 17.1 & - & 18.9 & - \\
    \bottomrule
    \end{tabular}
  \caption{Results from YOLACT~\cite{fsis_yolact} on our dataset and other few-shot instance segmentation datasets with respect to mAP and AP50. Results are presented for both instance segmentation and object detection.}
  \label{table:overall_yolact}
\end{table*}

\vspace{-1em}\paragraph{Localized Instance-Based Statistics.}
We next analyze the annotated object instances, characterizing each dataset's annotations as is for both segmentations and bounding boxes.  For every instance, we compute the following: 
\begin{enumerate}
    \itemsep0em 
    \item \textbf{Border rate}: ratio of the number of pixels in the object boundary touching the image border to the total number of pixels on the image border. An object touching the image border likely signifies that part of the object is outside the frame.
    \item \textbf{Center deviation}: ratio of the distance between the image center and the object center to the distance of the image center to its furthest pixel in the image. This indicates how far an object center's from the image center and so the traditional photographer's bias of centering visual content in the field of view.
    \item  \textbf{Image coverage}: ratio of the object area to the image area. This reveals the size of an object in the frame. 
    \item \textbf{Boundary complexity}: one minus the isoperimetric inequality, which is the ratio of the area of an instance to the length of its perimeter. A lower isoperimetric inequality value indicates a more complex boundary. This metric cannot be computed for bounding box annotations, so we exclud such datasets.
\end{enumerate} 

\noindent Results are shown in \textbf{Figure~\ref{fig:localization_analysis}}, with our dataset exhibiting a couple of important differences. \emph{First}, our dataset tends to have the largest \emph{border rate}, followed by VizWiz-FewShot (\textbf{Figure~\ref{fig:localization_analysis}(a)}). This finding aligns with prior work's quality assessments of BLV photographers' images, where inadequate framing including from out-of-frame areas was found in more than half of the (VizWiz) images~\cite{chiu2020assessing}.  \emph{Second}, our dataset tends to exhibit objects with the most \emph{image coverage} (\textbf{Figure~\ref{fig:localization_analysis}(b)}), thereby encouraging the development of algorithms that can handle an even wider range of object sizes. It is worth noting, that the positive correlation between \emph{border rate} and \emph{image coverage} likely arises for the same reason: closer proximity to the target object results in a larger object size within the image and so increased likelihood for the object to be partially out of frame.

%% file: 04-Benchmarking.tex
\section{Algorithm Benchmarking}
\label{sec:benchmark}
We now analyze modern methods' performance on our new dataset. We conduct analyses for both traditional few-shot localization algorithms as well as vision-language models (VLMs) in the zero-shot setting. 

\subsection{Few-shot Localization Algorithms}
\label{subsec:algo_fewshot}
We benchmarked the top-performing few-shot object detection (FSOD) and few-shot instance segmentation (FSIS) models for which publicly available code could be successfully deployed on modern GPUs: DeFRCN~\cite{fsod_DeFRCN} for FSOD and YOLACT~\cite{fsis_yolact} for both FSOD and FSIS. We adopted the hyperparameters and training pipelines from \cite{fsod_DeFRCN} for DeFRCN and \cite{tseng2022vizwiz} for YOLACT.  We trained models for base categories on the larger-scale VizWiz-FewShot~\cite{tseng2022vizwiz}, since those examples also originate from BLV individuals, and then fine-tuned them for novel categories on BIV-Priv-Seg. 

\vspace{-1em}\paragraph{Training and Evaluation Splits.}
The experiments were conducted using 4-fold cross-validation, where the 100 categories in VizWiz-FewShot were divided into four sets, with the combination of three sets (75 categories) serving as base categories in each fold. This matches prior work~\cite{tseng2022vizwiz} to enable fair comparison with VizWiz-FewShot~\cite{tseng2022vizwiz}.
%The 4-fold cross-validation ensures robust evaluation by training on three sets and validating on the remaining set in each fold. 

\vspace{-1em}\paragraph{Evaluation Metric.}
We evaluated the models using mean Average Precision (mAP), which is the average precision across a range of intersection over union (IoU) thresholds from 0.5 to 0.95 in steps of 0.05. This metric aligns with prior work~\cite{tseng2022vizwiz,fsod_DeFRCN,fsis_FAPIS,fsod_review1}. We also present results with $AP_{50}$, where only threshold $0.5$ is used to enable comparison with benchmarking on datasets such as Pascal VOC. 
%These metrics align with those used in prior works~\cite{tseng2022vizwiz,fsod_DeFRCN,fsis_FAPIS,fsod_review1}. While mAP highlights overall performance and robustness across a spectrum of localization accuracies, AP50 emphasizes the model's effectiveness at a specific, coarser alignment threshold.

\vspace{-1em}\paragraph{Baseline Datasets.}
We also analyzed each model's performance on other few-shot datasets: PASCAL-$5^{i}$\cite{fsss_oneshot}, COCO-20\textsuperscript{i}\cite{fsss_FeatureWeighting,fsis_siameseMRCNN}, and VizWiz-FewShot~\cite{tseng2022vizwiz}. 

\begin{table}[t!]
  \centering
  \footnotesize
  \vspace{-0.5em}
  \begin{tabular}{clcccc}
\toprule
Metric & Dataset~~~\textbackslash~~~$k = $ & 1 & 3 & 5 & 10 \\ 
\midrule 
& BIV-Priv-Seg (ours) & \textbf{12.7} & \textbf{19.4} & \textbf{23.8} & \textbf{29.7} \\
mAP &VizWiz-FewShot~\cite{tseng2022vizwiz} & 4.0 & 7.4 & 10.5 & 12.6 \\
&COCO-20\textsuperscript{i}~\cite{fsss_FeatureWeighting} & 9.3 & 14.8 & 16.1 & 18.5 \\
\midrule
& BIV-Priv-Seg (ours) & 21.7 & 30.3 & 36.4 & 44.7 \\
AP50 & VizWiz-FewShot~\cite{tseng2022vizwiz} & 6.9 & 12.8 & 17.6 & 23.5 \\
& PASCAL-5\textsuperscript{i}~\cite{fsss_oneshot} & \textbf{44.0} & \textbf{53.6} & \textbf{57.4} & \textbf{55.4} \\
    \bottomrule
    \end{tabular}%
    \caption{Results from DeFRCN~\cite{fsod_DeFRCN} on our dataset and other few-shot instance segmentation datasets with respect to two evaluation metrics: mAP and AP50.}
  \label{table:overall_defrcn}
\end{table}

\begin{figure*}[t!]
\includegraphics[width=1.0\textwidth]{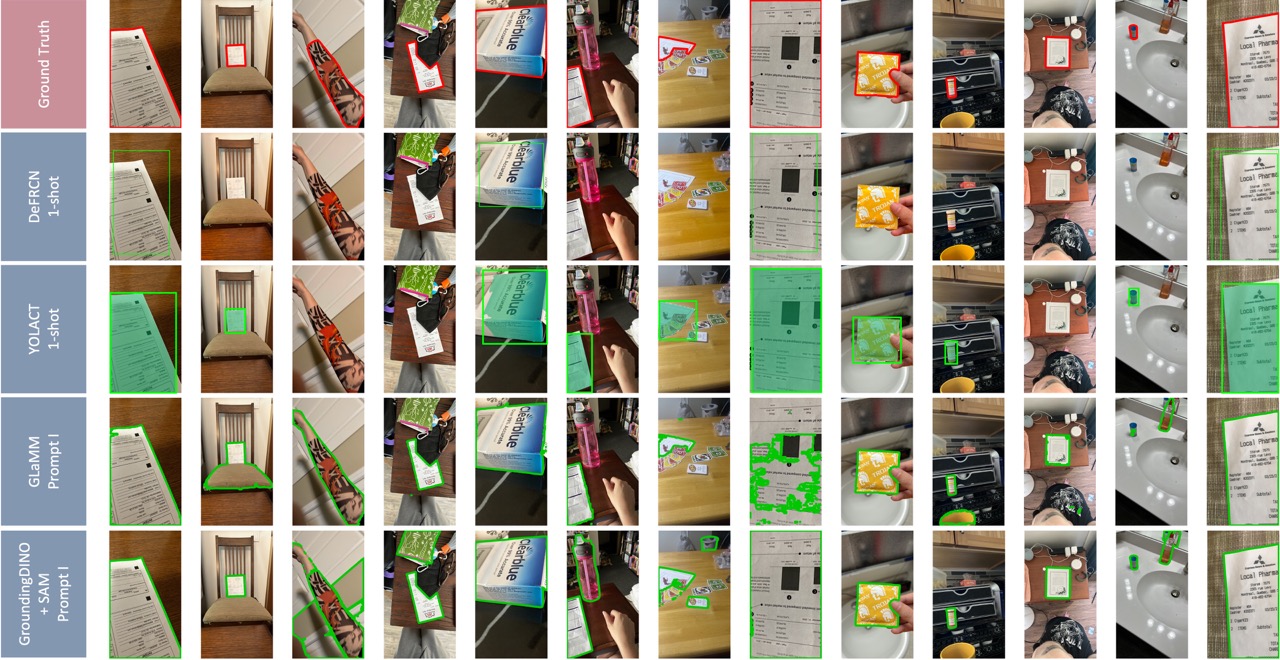}
% \vspace{-1.5em}
\caption{Qualitative results from DeFRCN in the 1-shot setting for object detection, YOLACT in the 1-shot setting for instance segmentation and object detection, and GLaMM and GroundingDINO+SAM using the prompt I scenario for the zero-shot setting.}
\label{fig:qualitative_results}
\end{figure*}

\vspace{-1em}\paragraph{Overall Performance.}
Results are in \textbf{Tables~\ref{table:overall_yolact}} and \textbf{\ref{table:overall_defrcn}}\footnote{The benchmarking results reported for DeFRCN performance on PASCAL-$5^{i}$ is presented in the format specified in~\cite{fsss_oneshot} with three data splits. To compare these results to the 4-fold cross-validation format, we average the scores of the three pre-defined splits.}.

Overall, we observe YOLACT performs poorly on all datasets with slightly better results on our dataset than that observed for the baseline datasets (\textbf{Table~\ref{table:overall_yolact}}).  For instance, the AP50 scores of the segmentations of our dataset are 24.46 and 38.53 for $k={1,5}$ compared to 19.0 and 21.7 of COCO-20\textsuperscript{i} and 12.03 and 24.99 of VizWiz-FewShot. We suspect this finding stems from our private objects having a notably higher prevalence of text and image coverage, as prior work~\cite{tseng2022vizwiz} has suggested such trends. 

Overall, we observe similar DeFRCN performance on our dataset to that observed for the baseline datasets (\textbf{Table~\ref{table:overall_defrcn}}).  With respect to the mAP scores, our dataset appears to be easier compared to VizWiz-FewShot and COCO-20\textsuperscript{i}, while it is harder than PASCAL-5\textsuperscript{i} when considering the AP50 scores.  A commonality across all the datasets is that the DeFRCN's performance is still quite poor, falling short of what might be practically useful in an end-user-facing tool.  Overall, our findings underscore the potential benefit of our dataset in helping advance modern few-shot learning algorithms.

Qualitative results from the 1-shot settings are shown for both models in \textbf{Figure~\ref{fig:qualitative_results}}. In these examples, we can see that YOLACT and DeFRCN tend to miss the same objects, while YOLACT locates some objects that DeFRCN misses.

%The mAP scores of our dataset are 20.04, 25.61, 31.06, and 44.42 for $k={1,3,5,10}$ compared to 9.3, 14.8, 16.1, and 18.5 of COCO-20\textsuperscript{i}~\cite{fsod_DeFRCN} and 4.03, 7.43, 10.52, and 12.61 of VizWiz-FewShot~\cite{tseng2022vizwiz}; the AP50 scores of our dataset are 21.70, 30.25, 36.41, and 44.69 for $k={1,3,5,10}$ compared to 44.03, 53.6, 57.43, and 55.37 of PASCAL-$5^{i}$~\cite{fsod_DeFRCN}. 

\vspace{-1em}\paragraph{Fine-Grained Analysis.}
We next perform fine-grained analysis of the models by dividing the test examples into subsets based on the following characteristics also employed by prior work~\cite{tseng2022vizwiz}: \emph{object size} and \emph{presence of text}. We categorize objects into small, medium, and large sizes, by sorting all instances in BIV-Priv-Seg based on their \emph{image coverage} and then equally divide them into three groups. Results are provided in \textbf{Table~\ref{table:FS_fine_grained_analysis}} with respect to mAP scores. The trends of AP50 scores are consistent with that of the mAP scores and so are excluded from the table.

For \emph{object size}, both model tend to perform better for larger objects, which is promising given that this size is a distinctly prominent feature in images from BLV photographers~\cite{reynolds2024salient,chen2022grounding}.  However, DeFRCN overall performs best on medium-sized objects rather than large-sized objects, suggesting a possibly slightly inferior match for BLV photographers. Both models perform the worst for smaller objects, reinforcing findings from prior work~\cite{fsod_FSRW, OD_review,tseng2022vizwiz}.

With respect to \emph{presence of text}, we observed that both models consistently perform better when objects have text. This reinforces findings in previous work~\cite{tseng2022vizwiz}, that text could serve as a valuable feature for models to locate objects in few-shot learning. This is particularly important for our privacy detection setting, since there is such a strong correlation between private content and the presence of text. %BIV-Priv-Seg includes a mixture of both text-based and non-text-based private information that challenges algorithms to develop a robust solution.

\begin{table}[t!]
  \centering
  \footnotesize
    \begin{tabular}{ccccccc}
    \toprule
&& \multicolumn{3}{c}{\textbf{\emph{Object Size}}} & \multicolumn{2}{c}{\textbf{\emph{Text}}} \\
\cmidrule(r){3-5} \cmidrule(r){6-7}
Model & $k$ & S. & M. & L. & Yes & No \\
\midrule
& $1$ & 12.41 & 22.60 & \textbf{26.53} & \textbf{24.56} & 11.24\\
YOLACT & $3$ &18.18 & 31.54 & \textbf{36.89} & \textbf{34.75} & 12.86  \\
(FSIS) & $5$ &22.90 & 36.97 & \textbf{44.39} & \textbf{42.32} & 18.31\\
&$10$ & 29.50 & 55.19 & \textbf{63.28} & \textbf{58.56} & 31.98 \\
\midrule
& $1$ & 8.70 & \textbf{15.50} & 13.33 & \textbf{15.63} & 4.95 \\
DeFRCN & $3$ & 11.71 & 23.21 & \textbf{23.29} & \textbf{23.76} & 8.08 \\
(FSOD) & $5$ & 15.73 & \textbf{26.49} & 25.11 & \textbf{28.13} & 11.45 \\
& $10$ & 17.89 & \textbf{34.67} & 30.32 & \textbf{34.58} & 15.02 \\
\bottomrule
\end{tabular}
\caption{Fine-grained analysis of the performance of FSIS and FSOD models on BIV-Priv-Seg presented in mAP.}
\label{table:FS_fine_grained_analysis}
\end{table}

\subsection{VLM Localization}
We next benchmarked top-performing, off-the-shelf vision language models that support visual grounding in a zero-shot setting in the following three ways:

\noindent \emph{\textbf{Prompt I. Privacy-agnostic (category specific)}}:
Each category's name is inserted into the prompt: \emph{``Segment it if there is a \{category name\} in the given image, otherwise say there is no \{category name\}."} 

\noindent \emph{\textbf{Prompt II. Privacy-aware (category agnostic)}}:
We use the following generic prompt \emph{``Segment it if there is a private object in the given image, otherwise say there is no private object."}, where we always use the generic phrase ``private object".  This prompt is intended to capture the model's understanding regarding what is private.

\noindent \emph{\textbf{Prompt III. Privacy-explained}}:
We provide context regarding what is private by providing the list of categories in our dataset in the prompt as follows: \emph{``Private objects include \{all category names\}. Segment it if there is a private object in the given image, otherwise say there is no private object."}

We intentionally designed these prompts to support when no target object is present, meaning that the model does not return a segmentation result. We achieved this by extending the prompts from previous work~\cite{hanoona2023GLaMM} from \emph{“Segment {something}”} to \emph{“Segment it if {something}, otherwise say there is no {something}”}. 

We tested four models, including three conversational VLMs and one text-prompted segmentation model. However, two of the conversational VLMs failed so terribly that we excluded them from detailed analysis: CogVLM~\cite{wang2023cogvlm} and GPT-4o coupled with Dall-E2~\cite{ramesh2022hierarchical}.  Specifically, CogVLM achieved a mAP score of 0 with prompts I, II, and III and GPT-4o consistently produced blank binary masks for our prompts and further tested variations (e.g., \emph{``Generate a binary mask that segments the \{category name\}."}). We also tested the conversational VLM, GLaMM~\cite{hanoona2023GLaMM}, and the text-prompted segmentation approach, GroundingDINO~\cite{liu2023grounding} coupled with Segment Anything (SAM)~\cite{kirillov2023segment}. Following the implementation in the GroundingDINO~\cite{liu2023grounding}, the prompts are simplified to single terms for GroundingDINO coupled with Segment Anything (SAM). Specifically, we adopted the designed prompts I and II and prompted the model as follows: (1) \emph{``\{category name\}"} and (2) \emph{``private object"}.
 
% The results are discussed below.

% as a baseline approach in Table~\ref{table:GroundingDINO_SAM_overall_seg}. We tested the text prompts (1) \emph{``\{category name\}"} and (2) \emph{``private object"}. This benchmarking helps establish a performance baseline for future studies aiming to improve text-prompted segmentation techniques. Results enhance the observations on GLaMM, where VLMs perform better with using specific category names as the prompt but struggle with the term \emph{``private object"}.
%These results highlight undertraining on the private categories introduced by our dataset or the lack of training for segmentation tasks, underscoring a vital area for future model development.

\vspace{-1em}\paragraph{Evaluation Method.}
We introduce a two-step evaluation method for handling when target objects are absent from an image. First, we categorize the images into a binary classification task with two classes based on whether private objects are found in the images. Then, we evaluate the localization results on the target-containing predictions.

\emph{Privacy classification:} For conversational algorithms (e.g., GLaMM), we evaluate the performance on two levels: (1) \textbf{Semantic level} prediction, which is based on the language used in the conversation, where positive responses such as \emph{“sure, here is the mask”} indicate the presence of a target, and negative responses such as \emph{“no, there is no...”} indicate the absence of a target; and (2) \textbf{Mask level} prediction, which is determined by whether any pixel in the generated mask has a positive prediction. The results are presented in accuracy, recall, and true negative rate. Accuracy shows the overall performance, the percentage of correct predictions among the target-containing images, and the percentage of correct predictions among the target-less images, respectively. For mask generation-only algorithms (e.g., GroundingDINO coupled with SAM), we only evaluate the results on \textbf{Mask level} prediction.

\emph{Segmentation:} For the images that are classified as private on both semantic and mask levels, we then perform localization. The mIoU and cIoU (class-weighted IoU)  metrics are used to evaluate the semantic segmentation results across all categories in each scenario. As most of the images in this dataset only contain one target object, we convert those segmentation results into instance segmentation format and evaluate the models in AP50 and Recall metrics. This two-step evaluation highlights the model's ability to identify whether there are private objects in the image and its localization ability.

\vspace{-1em}\paragraph{Privacy Classification Performance.}

The results of both the semantic level and mask level privacy classifications are shown in \textbf{Table~\ref{table:VLM_overall_cls}}. 

We observe three key findings. \emph{First}, on the \textbf{semantic level}, the GLaMM consistently generates positive responses across all scenarios, producing masks in every image. However, even then the model confirms the presence of a private object at the conversational level, some of the generated masks do not predict any pixels. This inconsistency suggests the model lacks a robust text-mask alignment mechanism. \emph{Second}, GLaMM's negative \textbf{mask level} predictions of are still quite inaccurate given the low true negative (TN) rates in prompt II and III. \emph{Finally}, GroundingDINO+SAM, unlike GLaMM, does not generate any negative responses. This underscores that models' struggle to identify when no target objects are in the images. We suspect that these issues stem from biases in the models' training data, where there is a lack of target-less samples.

\begin{table}[h!]
  \centering
  \footnotesize
    \begin{tabular}{lcccccc}
    \toprule
 & \multicolumn{3}{c}{Semantic level} &  \multicolumn{3}{c}{Mask level} \\
\cmidrule(r){2-4} \cmidrule(r){5-7} 

Prompt & Acc & R & TN & Acc & R & TN \\

\midrule
& \multicolumn{6}{c}{GLaMM} \\

\cmidrule(r){2-7}
Prompt I & 81.32 & 100.00 & \textbf{0.00} & 81.32 & 100.00 & \textbf{0.00}  \\
Prompt II & 81.32 & 100.00 & \textbf{0.00} & 69.26 & 83.97 & 5.21\\
Prompt III & 81.32 & 100.00 & \textbf{0.00} & 63.71 & 74.28 & 17.71 \\

\midrule
& \multicolumn{6}{c}{GroundingDINO + SAM} \\
\cmidrule(r){2-7}
Prompt I & - & - & - & 81.32 & 100.00 & \textbf{0.00}  \\
Prompt II & - & - & - & 81.32 & 100.00 & \textbf{0.00} \\
\bottomrule
\end{tabular}
\caption{Privacy classification performance of GLaMM and GroundingDINO coupled with SAM on BIV-Priv-Seg presented in accuracy (Acc), recall (R), and true negative rate (TN) as percentages. Results of GLaMM are evaluated on both semantic and mask levels, and those of GroundingDINO coupled with SAM are evaluated in mask level as there is no textual output. }
\label{table:VLM_overall_cls}
\end{table}

\begin{table}[t!]
  \centering
  \footnotesize
    \begin{tabular}{lccccc}
    \toprule
Evaluated Dataset & mIoU & cIoU & AP50 & Recall\\
% \cmidrule(r){2-5}
\midrule
& \multicolumn{4}{c}{GLaMM} \\
\cmidrule(r){2-5}
BIV-Priv-Seg (ours, prompt I) & \textbf{68.0} & 66.9 & \textbf{44.7} & \textbf{55.9} \\
BIV-Priv-Seg (ours, prompt II) & 11.4 & 11.4 & 3.1 & 7.8 \\
BIV-Priv-Seg (ours, prompt III) & 20.6 & 20.6 & 9.4 & 14.8 \\
VizWiz-FS~\cite{tseng2022vizwiz} (prompt I) & 15.6 & 13.4 & 2.2 & 5.6 \\
GranD (validation)~\cite{hanoona2023GLaMM} & 66.3 & - & 30.8 & 41.8 \\
GranD (test)~\cite{hanoona2023GLaMM} & 65.6 & - & 29.2 & 40.8\\
refCOCO (testA)~\cite{kazemzadeh2014referitgame} & - & \textbf{83.2} & - & - \\
% refCOCO (testB)~\cite{kazemzadeh2014referitgame} & - & 76.9 & - & - \\
refCOCO+ (testA)~\cite{kazemzadeh2014referitgame} & - & 78.7 & - & - \\
% refCOCO+ (testB)~\cite{kazemzadeh2014referitgame} & - & 64.6 & - & - \\
refCOCOg (test)~\cite{kazemzadeh2014referitgame} & - & 74.9 & - & - \\ 
\midrule
& \multicolumn{4}{c}{GroundingDINO + SAM} \\
\cmidrule(r){2-5}
BIV-Priv-Seg (ours, prompt I) & \textbf{73.2} & \textbf{70.2} & \textbf{60.2} & \textbf{69.6}\\
BIV-Priv-Seg (ours, prompt II) & 39.7 & 33.1 & 18.8 & 31.4 \\
VizWiz-FS (prompt I) & 39.0 & 40.2 & 17.4 & 35.1 \\
\bottomrule
\end{tabular}
\caption{Overall performance of the segmentation results of GLaMM and GroundingDINO coupled with SAM on BIV-Priv-Seg. BIV-Priv-Seg is evaluated with all prompts, and VizWiz-FewShot is evaluated with prompt I only as the dataset contains only non-private categories. GLaMM results for GranD, refCOCO, refCOCO+, and refCOCOg are adopted from \cite{hanoona2023GLaMM}.}
\label{table:VLM_overall_seg}
\end{table}

\vspace{-1em}\paragraph{Segmentation Performance.}
Segmentation results for the target-containing group are shown in \textbf{Table~\ref{table:VLM_overall_seg}}, with qualitative results in \textbf{Figure~\ref{fig:qualitative_results}}. We first compared the models' performance on our dataset to those on VizWiz-FewShot~\cite{tseng2022vizwiz}, GranD~\cite{hanoona2023GLaMM}, refCOCO, refCOCO+, and refCOCOg~\cite{kazemzadeh2014referitgame}. The results show that the performance of our dataset is comparable to that of existing datasets. Specifically, the performance of GLaMM on refCOCO, refCOCO+, and refCOCOg is better than BIV-Priv-Seg (prompt I), while its performance on GranD and VizWiz-FewShot is worse; and GroundingDINO coupled with SAM performs better on BIV-Priv-Seg than VizWiz-FewShot. This indicates that state-of-the-art VLMs have the potential to effectively ground private objects in images, achieving performance comparable to other datasets.

We then compared the performance of different prompts on BIV-Priv-Seg to explore the model's ability to comprehend privacy. Prompt I achieved the highest performance for both models, indicating that the model has the potential to locate private objects accurately when provided specific object categories. Conversely, prompt II had the worst performance. That may be in part because the models' notion of what is private may differ from what has been indicated by BLV individuals~\cite{stangl2020visual}, including because the concept can be highly subjective.  Interestingly, prompt III with GLaMM attempted to mitigate this semantic ambiguity by providing an explanation of privacy, yet it still performed poorly. Despite the initial definition of privacy in prompt III, the model failed to fully comprehend and execute the instructions. This suggest the model is highly sensitive to the sentence structure of prompts, which may reflect biases in its training data.

Overall, the performance of GroundingDINO coupled with SAM is better than GLaMM. We hypothesize that the conversational inputs for GLaMM add a challenging complexity for the grounding task, despite it potential benefit of offering end users more flexibility (e.g., prompt III).

We next conducted fine-grained analyses with respect to \emph{object size} and \emph{presence of text}, as done inSection 4.1. Also, because we observed from qualitative results that VLMs tend to segment the salient object, we evaluated based on whether the target object is the detected \emph{salient} object~\cite{saliency_kim2022revisiting} by checking if there was at least a $0.5$ threshold IoU between the segmentations. Results are shown in \textbf{Table~\ref{table:VLM_fine_grained_analysis}}. 

Results of \emph{object size} and \emph{presence of text} show performance increases from smaller to larger objects and yield better performance for objects with text. This reinforces findings from prior work~\cite{tseng2022vizwiz} for images taken by BLV individuals, underscoring that certain similarities between private and non-private objects can serve as potential opportunities in overcoming the natural scarcity of private data. We also observe consistently better performance for salient objects than non-salient objects, implying that modern foundation models leverage object saliency for grounding. 

%Following the implementation source \footnote{https://github.com/luca-medeiros/lang-segment-anything}, w

\begin{table}[t!]
  \centering
  \footnotesize
    \begin{tabular}{lccccccc}
    \toprule
& \multicolumn{2}{c}{\textbf{Saliency}} & \multicolumn{3}{c}{\textbf{Object Size}} & \multicolumn{2}{c}{\textbf{Text}} \\
\cmidrule(r){2-3} \cmidrule(r){4-6} \cmidrule(r){7-8} 
metric & Yes & No & S. & M. & L. & Yes & No \\
\midrule
& \multicolumn{7}{c}{GLaMM (prompt I)} \\
\cmidrule(r){2-8}
% \midrule
mIoU & \textbf{85.1} & 57.1 & 37.8 & 79.2 & \textbf{85.2} & \textbf{73.8} &  51.2 \\
cIoU & \textbf{84.2} & 56.1 & 36.9 & 78.1 & \textbf{85.2} &\textbf{72.7} & 50.4 \\
AP50 & \textbf{79.8} & 37.9 & 19.5 & 73.9 & \textbf{84.1} & \textbf{60.5} & 30.3 \\
Recall & \textbf{79.9} & 44.5 & 24.9 & 74.0 & \textbf{82.8} & \textbf{64.4} & 38.2 \\
\midrule

& \multicolumn{7}{c}{GroundingDINO + SAM (prompt I)} \\
% \midrule
\cmidrule(r){2-8}
mIoU & \textbf{91.1} & 67.8 & 57.3 & 79.9 & \textbf{91.1} & \textbf{82.5} &  59.4\\
cIoU & \textbf{92.5} & 65.7 & 57.3 & 77.8 & \textbf{92.9} &\textbf{82.1} & 53.2 \\
AP50 & \textbf{83.3} & 46.6 & 35.6 & 64.5 & \textbf{87.3} & \textbf{69.6} & 36.8 \\
Recall & \textbf{84.0} & 60.2 & 50.6 & 67.4 & \textbf{86.5} & \textbf{73.8} & 53.9 \\
\bottomrule
\end{tabular}
\caption{Fine-grained analysis of GLaMM and GroundingDINO coupled with SAM on BIV-Priv-Seg in prompt I scenarios.}
\label{table:VLM_fine_grained_analysis}
\end{table}

%% file: 05-Conclusion.tex
\section{Conclusions}
\label{sec:conclusions}
We introduce a new few-shot localization dataset challenge that contributes to developing privacy-preserving technologies for BLV users: BIV-Priv-Seg.  While there remain gaps between models' capabilities and privacy preservation needs, our public-sharing of the dataset offers a valuable foundation for future advancement of privacy-aware solutions.  Our results from model benchmarking underscore that a promising direction for future research is infusing privacy-aware capabilities into modern VLMs, particularly because BLV users already rely on such conversational agents (e.g., Be My AI~\cite{BeMyEyes:online}, Microsoft's Seeing AI~\cite{SeeingAI:online}).  More generally, we expect success on our dataset will facilitate progress for a broader number of scenarios facing similar challenges, such as for robotics and lifelogging.  

%Future research directions include reconciling VLMs' inconsistencies between text and mask predictions while also infusing them with the ability to recognize when requested objects are absent.

%Potential interactions include identifying private content in images, obfuscating the information, and providing detailed privacy status updates through a conversational interface. 

% We envision two important directions for future work. \emph{First}, is supporting personalized identification of private objects on one's own devices, where users train models to recognize their private objects in any new images or videos \emph{on-device}. A critical direction to make this idea a reality is developing such models to be compact so that they can run on resource-constrained devices. 